\UseRawInputEncoding
\documentclass[10pt,twocolumn,letterpaper]{article}

\usepackage{iccv}
\usepackage{times}
\usepackage{epsfig}
\usepackage{graphicx}
\usepackage{amsmath}
\usepackage{amssymb}
\usepackage{ulem}
\usepackage{multirow}
\usepackage{setspace}
\newcommand*{\affaddr}[1]{#1} 
\newcommand*{\affmark}[1][*]{\textsuperscript{#1}}
\newcommand*{\email}[1]{\texttt{#1}}
\usepackage[breaklinks=true,bookmarks=false]{hyperref}

\iccvfinalcopy 

\def\iccvPaperID{10356} 

\ificcvfinal\fi

\begin{document}

\title{FuseFormer: Fusing Fine-Grained Information \\ in Transformers for Video Inpainting}

\author{%
Rui Liu\affmark[\dag]\thanks{The first three authors contribute equally to this work.} \quad Hanming Deng\affmark[\ddag]\footnotemark[1] \quad Yangyi Huang\affmark[\ddag\S]\footnotemark[1] \quad Xiaoyu Shi\affmark[\dag] \quad Lewei Lu\affmark[\ddag] \\
Wenxiu Sun\affmark[\ddag$\sharp$] \quad Xiaogang Wang\affmark[\dag] \quad Jifeng Dai\affmark[\ddag] \quad Hongsheng Li\affmark[\dag\#] \\
\affaddr{\affmark[\dag]CUHK-SenseTime Joint Laboratory, The Chinese University of Hong Kong \quad \affmark[\ddag]SenseTime Research \\
\affmark[\S]Zhejiang University \quad \affmark[$\sharp$]Tetras.AI \quad \affmark[\#]School of CST, Xidian University}\\
\email{\{ruiliu@link, xiaoyushi@link, xgwang@ee, hsli@ee\}.cuhk.edu.hk} \\
\email{\{denghanming, huangyangyi, luotto, daijifeng\}@sensetime.com}\\
}

\maketitle
\ificcvfinal\thispagestyle{empty}\fi

\begin{abstract}
   Transformer, as a strong and flexible architecture for modelling long-range relations, has been widely explored in vision tasks.
   However, when used in video inpainting that requires fine-grained representation, existed method still suffers from yielding blurry edges in detail due to the hard patch splitting. Here we aim to tackle this problem by proposing FuseFormer, a Transformer model designed for video inpainting via fine-grained feature fusion based on novel Soft Split and Soft Composition operations.
   The soft split divides feature map into many patches with given overlapping interval. On the contrary, the soft composition operates by stitching different patches into a whole feature map where pixels in overlapping regions are summed up.
   These two modules are first used in tokenization before Transformer layers and de-tokenization after Transformer layers, for effective mapping between tokens and features.
   Therefore, sub-patch level information interaction is enabled for more effective feature propagation between neighboring patches, resulting in synthesizing vivid content for hole regions in videos.
   Moreover, in FuseFormer, we elaborately insert the soft composition and soft split into the feed-forward network, enabling the 1D linear layers to have the capability of modelling 2D structure. And, the sub-patch level feature fusion ability is further enhanced.
   In both quantitative and qualitative evaluations, our proposed FuseFormer surpasses state-of-the-art methods. We also conduct detailed analysis to examine its superiority.
   Code and pretrained models are available at \url{https://github.com/ruiliu-ai/FuseFormer}.
   \end{abstract}

\section{Introduction}
Transformer has recently gained increasing attention in various vision tasks such as classification~\cite{vit, yuan2021tokens}, object detection~\cite{detr, zhu2020deformable} and image generation~\cite{jiang2021transgan, hudson2021gansformer}.
Interestingly, Transformer is suitable to video inpainting, a vision task that depends on the information propagation between flowing pixels across frames to fill the spatiotemporal holes with plausible and coherent content in a video clip.

\begin{figure}[t]
    \centering
    \includegraphics[width=0.4\textwidth]{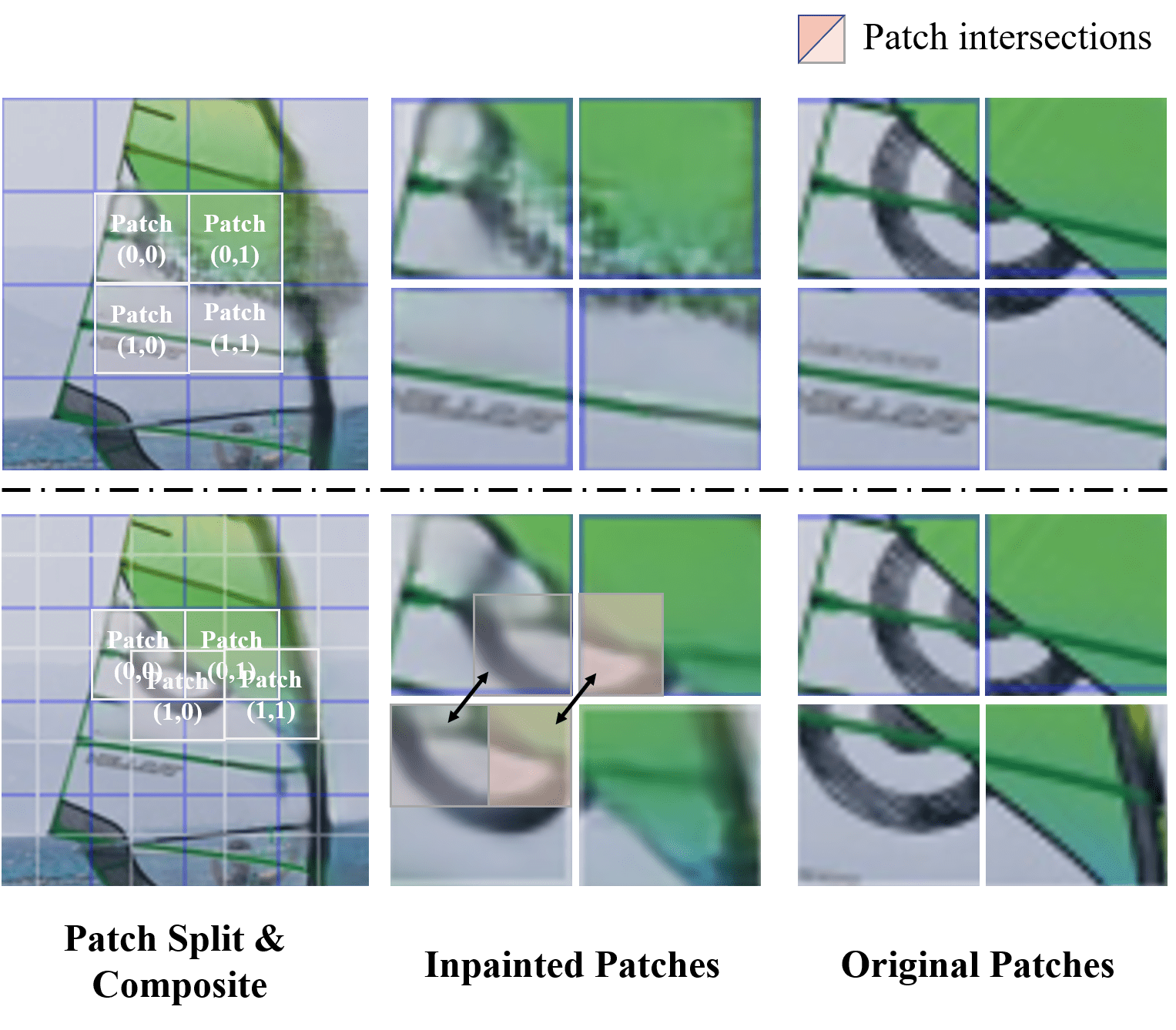}
    \caption{Illustration of different patch split/composition strategies for Transformer model. The top row shows hard split/composition, based on which the trained model generates rough inpainting results. The bottom row shows soft split/composition, based on which the trained model generates smooth results due to interaction of features between neighbor patches. Double arrow indicates the corresponding overlapped regions between adjacent patches.}
    \label{fig:motivation}
    \vspace{-1em}
\end{figure}

Spatial Temporal Transformer Net (STTN)~\cite{sttn} is the pioneer work for investigating the use of Transformer in video inpainting.
However, its multi-scale variant of self-attention intertwined with fully convolutional networks makes it hard to exploit rich experience from other Transformer models due to large structural differences.
On the other hand, recent Vision Transformer (ViT)~\cite{vit} demonstrates the strong capability of vanilla Transformer~\cite{attention} in vision recognition task.
These motivate us to build a Video inpainting Baseline with vanilla Transformer (ViB-T),
which differs from ViT in $2$ aspects: a) the tokens are embedded from patches of multiple frames instead of a single frame; b) a light convolutional encoder and decoder before and after Transformer block is exploited to relieve the computational burden caused by high resolution frames.
Experiment verifies that this simple baseline can reach competitive performance with STTN~\cite{sttn} under similar computation cost.

Nevertheless, similar to all existing patch-based Transformer models~\cite{vit, yuan2021tokens}, the hard split operation used in ViB-T makes it unable to effectively encode sub-token (sub-patch) level representations.
Since the attention score is calculated between different tokens, there is no direct sub-token level feature interaction.
For us, human beings, fragmenting an image into many non-overlapping patches poses a challenging task to composite them back into an original image with masked regions filled.
This is the same for deep learning systems: the lack of accurate sub-token level feature interaction can lead to inconsistent content between neighboring patches. As shown in Fig.\ref{fig:motivation}, to accurately rebuild the black circle on the canvas, every token corresponding to an image patch has to understand not only the patch level information but also sub-patch level information.
As a result, in order to fully unleash the power of Transformers in video inpainting tasks, an improved patch splitting manner and a better sub-token level feature fusion mechanism to maintain pixel level-feature accuracy is in demand.

To achieve this goal, we propose a Soft Split (SS) module as well as its corresponding Soft Composition (SC) module.
Built upon the simple and straightforward ViB-T baseline model, we propose to softly split images into patches with overlapping regions and correspondingly, to softly composite these overlapped patches back to images.
Specifically, in the soft split module, we exploit an \textit{unfold} operation with kernel size greater than stride to softly split the input image into overlapping $2D$ patches and are flattened as $1D$ tokens.
On the contrary, in the soft composition module, tokens are reshaped to $2D$ patches maintaining their original sizes, and then each pixel is registered to its original spatial location according to the kernel size and stride used in soft split module.
During this process, features of the pixels located in the overlapping area are fused from multiple overlapping neighboring patches' corresponding areas, thus providing sub-token level feature fusion.
We design a baseline ViB-T model equipped with the Soft Split and Soft Composition modules as ViB-S where S stands for \textit{soft} operations.
And we find that the ViB-S model easily surpasses the state-of-the-art video inpainting model STTN \cite{sttn} with minimum extra computation cost.

Finally, we propose a Fusion Feed Forward Network (F3N) to replace the two-layer MLPs in the standard Transformer model, which is dubbed as FuseFormer, to further improve its sub-token fusion ability for learning fine-grained feature, yet without extra parameters.
In the F3N, between the two fully-connected layers, we reshape each $1D$ token back to $2D$ patch with its original spatial shape and then softly composite them to be a whole image.
The overlapping features of pixel at overlapping regions would sum up the corresponding value from all neighboring patches for further fine-grained feature fusion. Then the patches are softly split and flattened into $1D$ vectors, which are fed to the second MLP.
In this way, sub-token segment corresponding to the same pixel location are matched and registered without extra learnable parameters, and information of the same pixel location from different patches are aggregated. Subsequently, our FuseFormer model consisting of F3N even surpasses our strong baseline ViB-S by a significant margin, both qualitatively and quantitatively.

Based on these novel designs, our proposed FuseFormer network achieves effective and efficient performance in video restoration and object removal.
We testify the superiority of the proposed model to other state-of-the-art video inpainting approaches by thorough qualitative and quantitative comparisons. We further conduct ablation study to show how each component of our model benefits the inpainting performance.

In summary, our contributions are three-fold:

\begin{enumerate}
    \item We first propose a simple yet strong Transformer baseline for video inpainting, and propose a soft split and composition method to boost its performance.
    \item Based on the proposed strong baseline and novel soft operations, we propose FuseFormer, a sub-token fusion enabled Transformer model with no extra parameters.
    \item Extensive experiments demonstrate the superiority of FuseFormer over state-of-the-art approaches in video inpainting, both qualitatively and quantitatively.
\end{enumerate}


\section{Related work}
\noindent \textbf{Image Inpainting}. In traditional image inpainting, the target holes are usually filled by sampling and pasting the known textures and significant progress has been made on this type of image inpainting approach~\cite{10.1145/344779.344972, 10.1109/TIP.2003.815261, ImageMelding12, Efros99texturesynthesis, 10.1145/383259.383296}.  PatchMatch~\cite{patchmatch} proposes to fill the missing region by searching the patches outside the hole based on the approximate nearest neighbor algorithm, which is finally served as a commercial product.

With the rise of deep neural network~\cite{alexnet, He2015} and generative adversarial network~\cite{gan}, some works investigated on building an end-to-end deep neural network for image inpainting task with the auxiliary discriminator and adversarial loss~\cite{pathakCVPR16context, IizukaSIGGRAPH2017}.
After that, DeepFill propose to use a contextual attention for filling target holes by propagating the feature outside the region~\cite{yu2018generative}. Then Liu \textit{et al.} and Yu \textit{et al.} apply partial convolution~\cite{liu2018partialinpainting} and gated convolution~\cite{yu2018free} to make vanilla convolution kernels aware of given mask guidance respectively, so as to complete free-form image inpainting.

\noindent \textbf{Video Inpainting}.
Building upon patch-based image inpainting,
Newson \textit{et al.} extend PatchMatch algorithm~\cite{patchmatch} to video
for further modelling the temporal dependencies and accelerating the process of patch matching~\cite{Newson2014:SIIMS-VideoInpainting}.
Strobel \textit{et al.}~\cite{Strobel2014FlowAC} introduce an accurate motion field estimation for capturing object movement.
Huang \textit{et al.} perform an alternate optimization on $3$ steps including patch search, color completion and motion field estimation and obtain successful video completion performance~\cite{huang2016videocompletion}.

Deep learning also boosts the performance of video inpainting. Wang \textit{et al.} proposes a groundbreaking deep neural network that combines $2$D and $3$D convolution seamlessly for completing missing contents in video~\cite{wang2019aaai}. Kim \textit{et al.} propose a recurrent neural network to cumulatively aggregate temporal features through traversing all video sequences~\cite{vinet}. Xu \textit{et al.} use existing flow extraction tools to obtain robust optical flow and then warp the regions from reference frames to fill the hole in target frame~\cite{dfvi}. Lee \textit{et al.} propose a copy-and-paste network that learns to copy corresponding contents in reference frames and paste them to fill the holes in the target frame~\cite{cap}. Chang \textit{et al.} develop a learnable Gated Temporal Shift Module and adapt gated convolution\cite{yu2018free} to a $3$D version for performing free-form video inpainting~\cite{lgtsm, chang2019free}. Zhang \textit{et al.} adopts internal learning to train one-size-fits-all model for different given videos~\cite{zhang2019internal}. Hu \textit{et al.} propose a region proposal-based strategy for picking up best inpainted result from many participants~\cite{hu2020proposal}.
Recently attention mechanisms are adopted to further promote both realism and temporal consistency via capturing long-range correspondences in video sequences.
Temporally-consistent appearance is implicitly learned and propagated to the target frame with a frame-level attention~\cite{opn} and dynamic long-term context aggregation module \cite{shortlongterm}.

\noindent \textbf{Transformers in Vision}.
Transformers are firstly proposed in 2017~\cite{attention} and gradually dominated natural language processing models~\cite{devlin2018pretraining, nlppre, liu2019roberta}. A Transformer block basically consists of a multi-head attention module for modelling long-range correspondence of the input vector and a multi-layer perceptron for fusing and refining the feature representation. In computer vision, it has been adapted to various tasks such image classification~\cite{vit, yuan2021tokens}, object detection and segmentation~\cite{detr, zhu2020deformable, transcluster, SMCA}, image generation~\cite{jiang2021transgan, hudson2021gansformer}, video segmentation~\cite{wang2020end}, video captioning~\cite{zhou2018end} and so on in past two years.

As far as our knowledge concerns, STTN~\cite{sttn} is the only work for investigating the use of Transformer in video inpainting and propose to learn a deep generative Transformer model along spatial-temporal dimension.
It roughly splits frames into non-overlapped patches with certain given patch size and then feeds the obtained spatiotemporal patches into a stack of Transformer encoder blocks for thorough spatiotemporal propagation.
However, it suffers from capturing local texture like edges and lines and modelling the arbitrary pixel flowing.
In this work, we propose a novel Transformer-based video inpainting framework endorsed by $2$ carefully-designed soft operations, which improve the performance on both video restoration and object removal and make the inference much faster as well.

\begin{figure*}[t]
    \centering
    \includegraphics[width=0.98\textwidth]{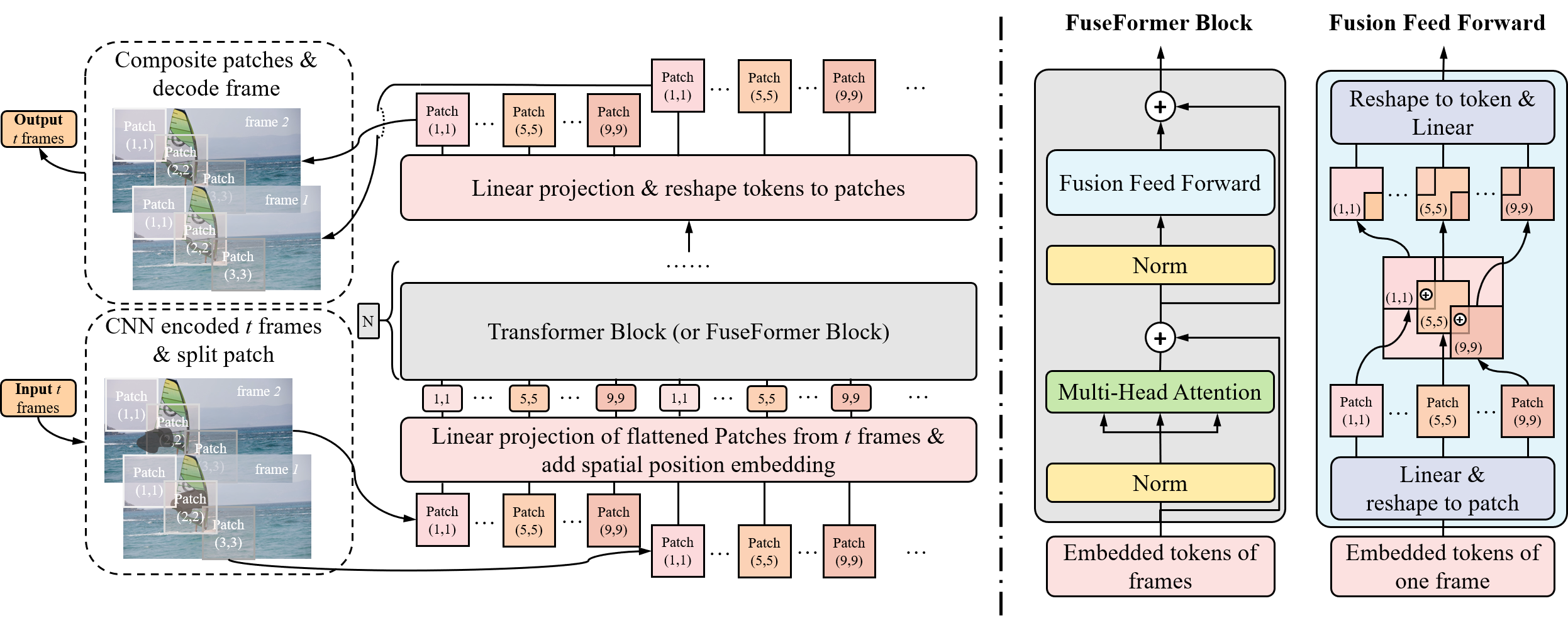}
    \caption{Illustrations of our proposed FuseFormer. On the left is our proposed video inpainting pipeline with Transformers. On the right is our proposed FuseFormer block and Fusion Feed Forward Network (F3N). The tuple indicates the counting number of patch along spatial dimension. }
    \label{fig:pipeline}
    \vspace{-1em}
\end{figure*}

\section{Method}
In this section we introduce our FuseFormer model for video inpainting. We start by proposing a simple Transformer baseline, named ViB-T (Video inpainting Baseline with vanilla Transformer), then we introduce our novel designs step by step by first introducing our Soft Split (SS) and Soft Composition (SC) technique, which boost the performance of ViB-T. We term ViB-T with SS and SC as ViB-S. Finally, build upon ViB-S, we introduce FuseFormer, a fine-grained vision Transformer block whose regular feed forward network is replaced with fusion feed forward network, and term the final model as ViF (Video inpainting with FuseFormer).

\subsection{Video inpainting Baseline with Transformer}
We start by proposing a straightforward baseline model ViB-T for directly deploying patch-based Transformer in video inpainting without complex modifications. It consists of three parts: a) a convolutional encoder and a corresponding decoder; b) a stack of Transformer blocks between the encoder and decoder; and c) a pair of patch-to-token and token-to-patch module. The patch-to-token module locates between the convolutional encoder and the first Transformer block, and token-to-patch locates between the last Transformer block and the convolutional decoder. Different from STTN~\cite{sttn}, this baseline model's Transformer block is the same as standard Transformer~\cite{attention} where there is neither the scheme of multi-scale frames for different multi-head self-attention nor using \(3 \times 3\) convolution to replace linear layers in feed forward network. Patches are hard split from feature map and linearly embedded to feature vectors with much lower channel dimension, which is more computationally friendly for following processing.

As shown in Fig.~\ref{fig:pipeline}, given corrupted video frames \(f_i\in\mathbb{R}^{h\times w \times 3}, i \in [0,t)\), it would work as follows:

First, it encodes video frames with a CNN encoder, obtaining \(c\) channel convolutional feature maps of frames \(\boldsymbol{X}_i \in \mathbb{R}^{h/4 \times w/4 \times c}, i \in [0,t)\), and each \(\boldsymbol{X}\) is split into \(k \times k\) smaller patches with stride \(s\). Then all patches are linearly embedded into tokens \(\boldsymbol{Z} \in \mathbb{R}^{(t\cdot n) \times d}\), where \(n\) is the number of tokens in one image and \(d\) is the token channel.

Second, \(\boldsymbol{Z}\) is fed into standard Transformer blocks for spatial-temporal information propagation, resulting in refined tokens \(\boldsymbol{\tilde{Z}}  \in \mathbb{R}^{(t\cdot n) \times d}\).

Third, each refined token \(\boldsymbol{\tilde{z}}_i \in \mathbb{R}^d, i \in [0, n\cdot t)\) from \(\boldsymbol{\tilde{Z}}\) is linearly transformed to \(k \cdot k \cdot c\) channel vector and reshaped to patch shape \(k \times k \times c\). All the resulting patches are registered back to its original frame's location pixel by pixel, obtaining feature maps \(\boldsymbol{\tilde{X}}_i \in \mathbb{R}^{h/4 \times w/4 \times c}, i \in [0,t)\). This re-composited feature map is of the same size as the feature map input to the first  Transformer block.

Finally, the re-composited feature maps \(\boldsymbol{\tilde{X}}\) are decoded with a couple of deconvolution layers to output the inpainted video frames \(\tilde{f}_i\in\mathbb{R}^{h\times w \times 3}, i \in [0,t)\) with original size.

For the baseline model ViB-T, we set kernel size equal to the stride in patch splitting. As a starting point, this simple model already has competitive performance with STTN \cite{sttn} but with faster inference speed and fewer parameters (refer to appendix C).

The key of our proposed method is the sub-token level fine-grained feature fusion, which is realized by the newly-proposed Soft Split (SS) and Soft Composite (SC) processing, it enables precise sub-token level fusion between neighboring patches. In the following section, we will first introduce the SS and SC modules, based on which we introduce our proposed FuseFormer in section 3.3.

\subsection{Soft Split (SS) and Soft Composite (SC)}
Different from STTN~\cite{sttn} that roughly split frames into patches without overlapping region, here we propose to softly split each frame into overlapped patches and then softly composite them back, by using an unfold and fold operator with patch size \(k\) being greater than patch stride \(s\). When compositing patches back to its original spatial shape, we add up feature values at each overlapping spatial location of neighboring patches.

\begin{figure}[t]
    \centering
    \includegraphics[width=0.48\textwidth]{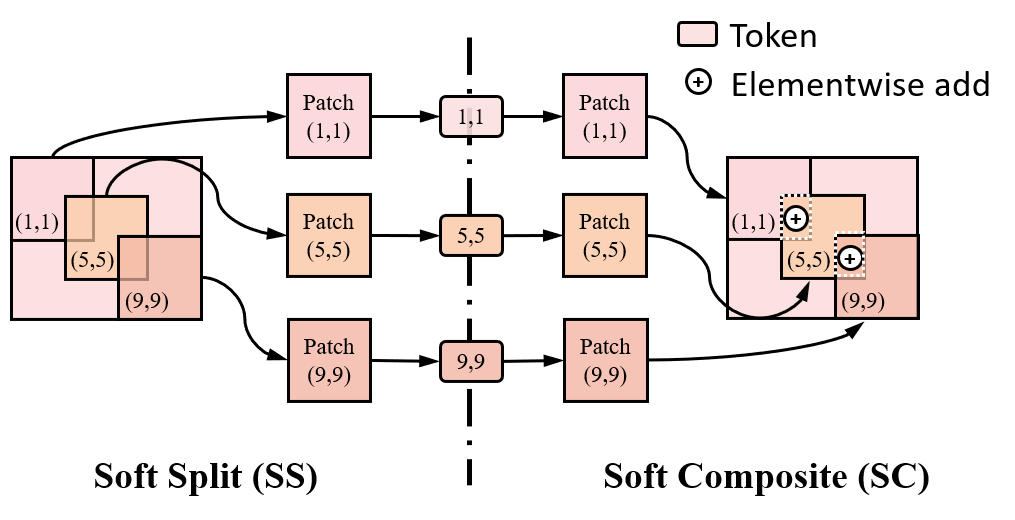}
    \caption{The illustration of Soft Split (SS) and Soft Composite (SC) module.}
    \label{fig:p2tt2p}
    \vspace{-1em}
\end{figure}

\paragraph{Soft Split (SS).} As shown in Fig.~\ref{fig:p2tt2p}, it softly split each feature map into overlapped patches of size \(k \times k\) with stride \(s < k\), and flattened to a one-dimensional token, which is similar to the image spliting strategy in T2T-ViT \cite{yuan2021tokens}. The number of tokens is then
\begin{equation}
    \label{equ:ntoken}
    n=\lfloor\frac{h + 2\cdot p-k}{s}+1\rfloor \times \lfloor\frac{w + 2\cdot p-k}{s}+1\rfloor,
\end{equation}
where \(p\) is the padding size.

\paragraph{Soft Composite (SC).} The SC operator composites the softly split \(n\) patches by their original spatial location and form a new feature map with the same \(h\) and \(w\) as original feature map size. However, due to the existence of overlapping area, the SC operator sums up pixel values that overlapped on the same spatial location, as shown in Fig.~\ref{fig:p2tt2p}.

This design of soft split and composition lays foundation for our final FuseFormer, as when softly compositing patches back to its original position after Transformer processing, the overlapped position aggregated a piece of information from different tokens, contributing to smoother patch boundaries and enlarges its receptive field by fusing information from neighboring patches. As our experiment shows, the baseline model equipped with these two operators, dubbed as ViB-S, have already surpassed the state-of-the-art video inpainting performance reached by STTN~\cite{sttn}.

\subsection{FuseFormer}
A FuseFormer block is the same to standard Transformer block except that feed forward network is replaced with our proposed Fusion Feed Forward Network (F3N). Given input patch tokens \(\boldsymbol{Z}_{l}\) at $l$-th stack where \(l\in [0,L)\), $L$ is the stacking number of FuseFormer blocks, a FuseFormer block can be formulated as:
\begin{align}
    \label{equ:msa}&\boldsymbol{Z'}_{l} = \text{MSA}(\text{LN}_1(\boldsymbol{Z}_{l - 1})) + \boldsymbol{Z}_{l}, \\
    \label{equ:ffn}&\boldsymbol{Z}_{l+1} =  \text{F3N}(\text{LN}_2(\boldsymbol{Z'}_{l})) + \boldsymbol{Z'}_{l},
\end{align}
where the MSA and LN respectively denote standard multi-head self-attention and layer normalization in Transformers~\cite{attention} and our key difference from other Transformers lies in the newly-proposed Fusion Feed Forward Network (F3N).

\paragraph{Fusion Feed Forward Network (F3N).} F3N brings no extra parameter into the standard feed forward net and the difference is that F3N inserts a SC and a SS operation between the two layer of MLPs. For clear formulation, we let $\boldsymbol{F'}=\text{F3N}(\boldsymbol{F})=\text{F3N}(\text{LN}_2(\boldsymbol{{Z'}_l}))$ where $\boldsymbol{F},\boldsymbol{F'} \in \mathbb{R}^{tn \times d}$ and the mapping functions are the same to Equ.~\ref{equ:ffn}. Let \(\boldsymbol{f}_i, \boldsymbol{f'}_i \) be the token vectors from \(\boldsymbol{F}, \boldsymbol{F'}\) where $i \in [0, t\cdot n)$, so the F3N can be formulated as
\begin{align}
    \label{equ:f3n0}&\boldsymbol{p}_i = \text{MLP}_1(\boldsymbol{f}_i), & i\in[0,t\cdot n) \\
    \label{equ:f3n1}&\boldsymbol{A}_j = \text{SC}(\boldsymbol{p}_{j, 0},..., \boldsymbol{p}_{j, n-1}), & j\in[0,t) \\
    \label{equ:f3n2}&\boldsymbol{p'}_{j, 0},..., \boldsymbol{p'}_{j, n-1} = \text{SS}(\boldsymbol{A}_j), & j\in[0,t) \\
    \label{equ:f3n3}&\boldsymbol{f'}_i = \text{MLP}_2(\boldsymbol{p'}_i), & i\in[0,t\cdot n)
\end{align}
where $\text{MLP}_1$ and $\text{MLP}_2$ denote the vanilla multi-layer perceptron. \text{SC} denotes soft composition for composing those $1$-D vectors $\boldsymbol{p}_{j, 0},..., \boldsymbol{p}_{j, n-1}$ to a $2$-D feature map $\boldsymbol{A}_j$ and SS denotes the soft split for splitting $\boldsymbol{A}_j$ into $1$-D vectors $\boldsymbol{p'}_{j, 0},..., \boldsymbol{p'}_{j, n-1}$. Note that there is a feature fusion processing during the mapping $\boldsymbol{p'}_i = \text{SS}( \text{SC}(\boldsymbol{p}_i))$.

Besides the introduction of soft composition and soft split module, there is another difference between F3N and FFN. In FFN, the input and output channel of \(\text{MLP}_1\) and \(\text{MLP}_2\) are \((4 \cdot d, d)\) and \((d, 4 \cdot d)\), respectively. On the contrary, in F3N, we change the input and output channel of the two MLPs to \(
(d, k^2\cdot c')\) and \((k^2\cdot c', d)\), where \(c'=10\cdot\lfloor 4\cdot d / (10\cdot k^2)\rfloor\), which aims to ensure the intermediate feature vectors are able to be reshaped to feature $2$-D maps.


For each soft composition module in F3N, different pixel locations may correspond to various number of overlapping patches, which leads to large variance on pixel value. Meanwhile, the spatial location of the reshaped patch is actually mixed up after passing through the $\text{MLP}_1$. Therefore, we introduce a normalization for Equ.~\ref{equ:f3n1}. Let \(\boldsymbol{1} \in \mathbb{R}^{n \times (k^2\cdot c')}\) be the vectors where all elements' value are $1$, so the normalized SC can be formulated as:
\begin{equation}
    \label{equ:meanfold}\boldsymbol{\tilde{A}}_j = \frac{\text{SC}(\boldsymbol{p}_{j, 0},..., \boldsymbol{p}_{j, n-1})}{\text{SC}(\boldsymbol{1})}, j\in[0,t)
\end{equation}

\subsection{Training Objective}
We train our network by minimizing the following loss:
\begin{equation}
\label{eq:final}
\begin{aligned}
  \mathcal{L} = \lambda_{\text{R}} \cdot \mathcal{L}_{\text{R}} + \lambda_{\text{adv}} \cdot \mathcal{L}_{\text{adv}},
\end{aligned}
\end{equation}
where $\mathcal{L}_{\text{R}}$ is the reconstruction loss for all pixels, $\mathcal{L}_{\text{adv}}$ is the adversarial loss from GAN~\cite{gan}, $\lambda_{\text{R}}$ and $\lambda_{\text{adv}}$ weigh the importance of different loss functions. For reconstruction loss, $L1$ loss is utilized for measuring the distance between synthesized video $\tilde{\mathbf{Y}}$ and original one $\mathbf{Y}$. It can be formulated as
\begin{equation}
\label{eq:hole}
\begin{aligned}
  \mathcal{L}_\text{R} = \|(\tilde{\mathbf{Y}} - \mathbf{Y})\|_1
\end{aligned}
\end{equation}

In addition, following~\cite{sttn}, we also adopt a discriminator $D$ for assisting training the FuseFormer generator, in order to obtain a better synthesis realism and temporal consistency. This discriminator takes both real videos and synthesized ones as input and outputs a scalar ranging in $[0,1]$ where $0$ indicates fake and $1$ indicates true. It is trained toward the direction that all the synthesized videos could be distinguished from real ones. The FuseFormer generator is trained towards an opposite direction where it generates videos that can not be told by $D$ anymore.
The loss function for $D$ is formulated as
\begin{equation}
\label{eq:advD}
\begin{aligned}
  \mathcal{L}_{D} =  \mathbb{E}_{\mathbf{Y}} \left[ \log{D(\mathbf{Y}} \right]
  +  \mathbb{E}_{\tilde{\mathbf{Y}}} \left[ \log{ (1 - D(\tilde{\mathbf{Y}})) } \right]
\end{aligned}
\end{equation}

And the loss function for the FuseFormer generator is
\begin{equation}
\label{eq:advG}
\begin{aligned}
  \mathcal{L}_{\text{adv}} = \mathbb{E}_{\tilde{\mathbf{Y}}} \left[ \log{ D(\tilde{\mathbf{Y}}) } \right]
\end{aligned}
\end{equation}

\section{Experiments}
\begin{figure}[t]
    \centering
    \includegraphics[width=0.48\textwidth]{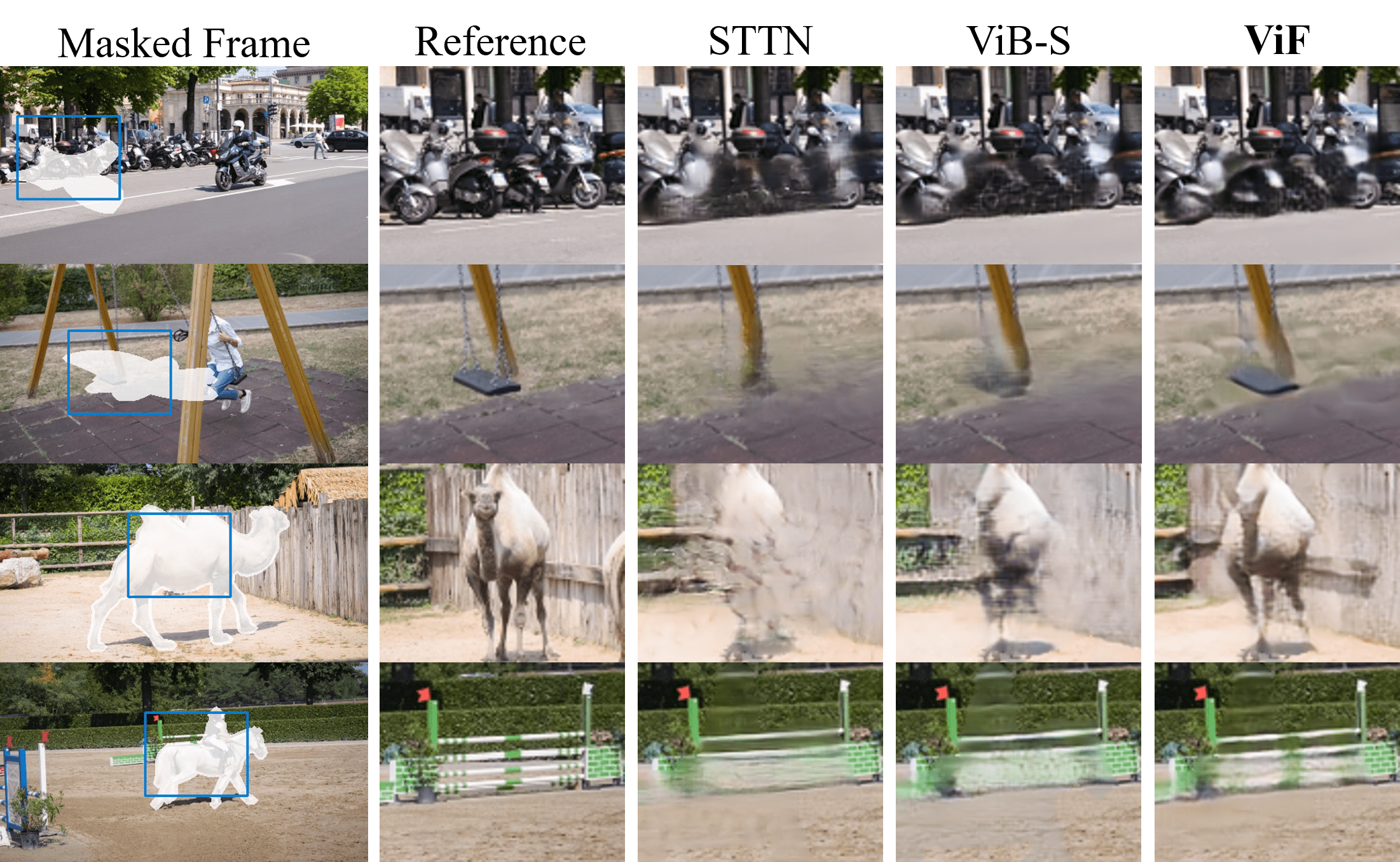}
    \caption{Qualitative results of our proposed ViB-S and ViF. Reference denotes masked object found in the same video. Compared to STTN, with soft patch split/composition, our ViB-S can better handle detail information. When replacing Transformer block in ViB-S with FuseFormer, ViF excels at recovering details and heavily occluded objects.}
    \label{fig:f3n_vis}
\end{figure}


\subsection{Implementation details}
\noindent \textbf{Dataset.} Following previous works~\cite{sttn,cap}, we choose $2$ video object segmentation datasets for training and evaluation.
\textit{YouTube-VOS}~\cite{youtubevos} contains $3,471$, $474$ and $508$ video clips in training, validation and test set, respectively.
\textit{DAVIS}~\cite{Perazzi2016}, short for Densely Annotated Video Segmentation, contains $150$ video sequences in various scenes.
Following STTN~\cite{sttn}, a test set including $60$ video clips is split from the whole dataset for fair comparison with other approaches. We do not use this dataset for training.

\noindent \textbf{Network and training.} We use $8$ stacks of Transformer (FuseFormer) layers in our ViB-T, ViB-S and ViF models, whose token dimension is $512$. For ViF, the token is expanded to $1960$ instead of $2048$ for patch reshape compatibility. Other network structures including the CNN encoder, decoder and discriminator are the same as STTN~\cite{sttn}, except that we insert several convolutional layers between encoder and the first Transformer block to compensate for aggressive channel reduction in patch tokenization.
Note that different from STTN~\cite{sttn}, we do not finetune our model on DAVIS training set and the same checkpoint is used for evaluation on both YouTube-VOS test set and DAVIS test set.
In all our ablations, we train our model with Adam optimizer \cite{adam} for 250k iterations. At each iteration, $5$ random frames from one video is sampled on each GPU and $8$ GPU is utilized. The initial learning rate is $0.01$ and is reduced by factor of $10$ at 200k iteration. For our fair comparison with state-of-the-art models, we train our best model for $500$k iterations, and the learning rate is reduced at $400$k and $450$k iterations respectively.

\begin{table}[t]
\centering
 \begin{tabular}{l c c c c}
 \hline
 Model & Patch Size & Overlap & PSNR $\uparrow$ & SSIM $\uparrow$ \\ [0.5ex]
 \hline\hline
 STTN \cite{sttn} & (5,9)$^*$ & no &30.67 &0.9560 \\
 \hline
 ViB-T & (3,3) & no & 30.68 & 0.9569  \\
 ViB-T & (5,5) & no & 30.56 & 0.9563  \\
 ViB-T & (7,7) & no & 30.50 & 0.9559 \\
  \hline
 ViB-S$^{\rhd}$ & (7,7) & yes & 30.74 & 0.9577 \\
 ViB-S$^{\lhd}$ & (7,7) & yes & 30.99 & 0.9597 \\
 \hline
 \multirow{2}{*}{ViB-S} & (5,5) & yes & 30.91 &0.9588 \\
  & (7,7) & yes & 31.02 & 0.9598\\
  \hline
 ViF$^{\dagger}$ & (7,7) & yes & 31.72 & 0.9654 \\
 ViF & (7,7) & yes &  \textbf{31.87} & \textbf{0.9662} \\
 \hline
 \end{tabular}
 \label{tab:kernelsize}
 \caption{Evaluation of our proposed SS, SC module and FuseFormer. All models except STTN use patch stride of $3$. ViB-S$^{\rhd}$ and ViB-S$^{\lhd}$ denotes using only SC or SS respectively. ViF$^\dagger$ denote using F3N without normalizing in Equ.\ref{equ:meanfold} and ViF denote using F3N with normalizing. $^*$: STTN uses multi-scale patch sizes and refer to~\cite{sttn} for more details. }
\end{table}

\begin{figure*}[t]
    \centering
    \includegraphics[width=0.98\textwidth]{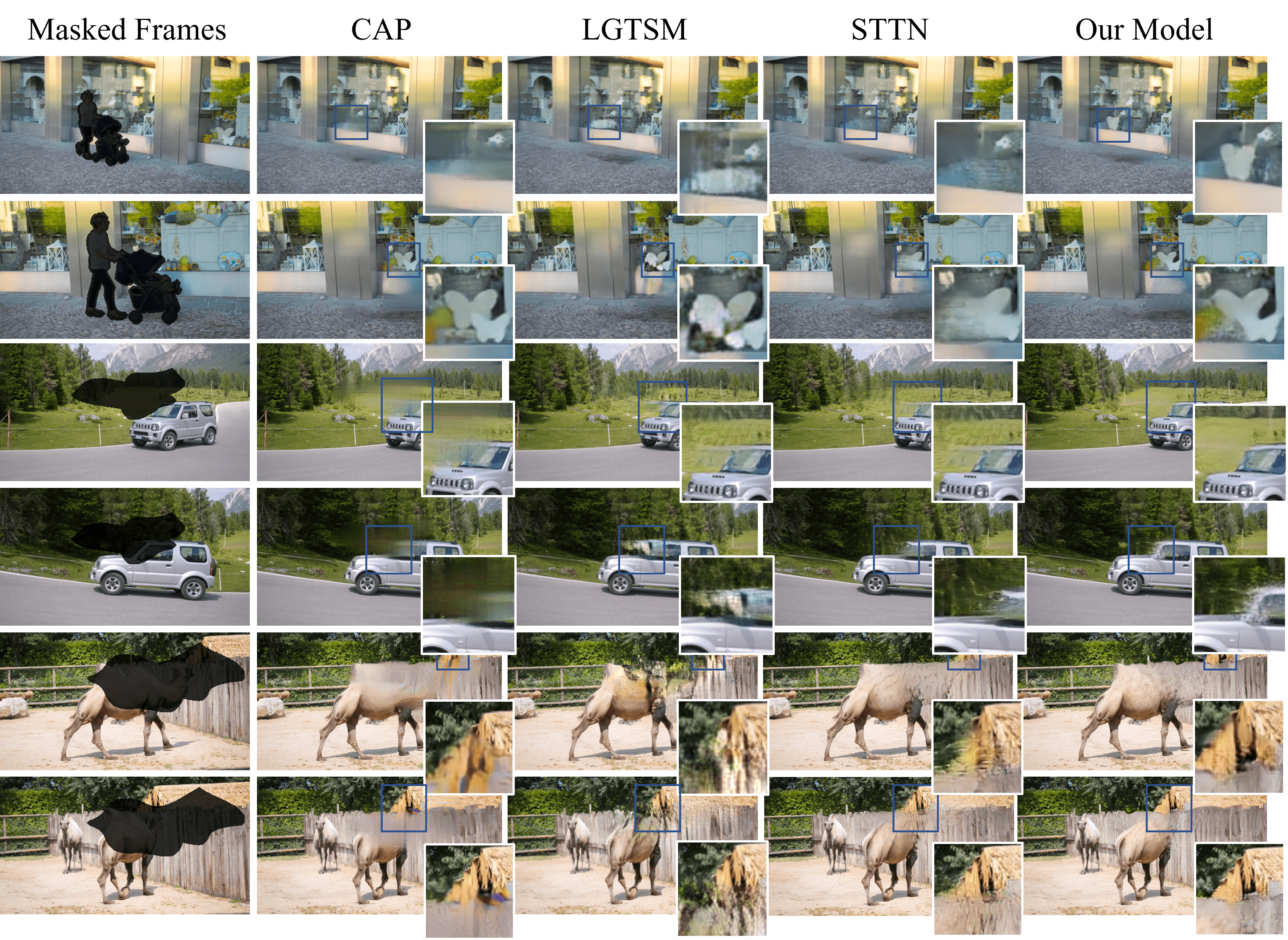}
    \caption{Qualitative comparison with other methods.}
    \label{fig:final_vis}
\end{figure*}

\begin{table*}[t]
\begin{center}

\begin{tabular}{l|c|c|c|l||c|c|c|l}
\hline
& \multicolumn{8}{c}{Accuracy}  \\
\hline
& \multicolumn{4}{c||}{YouTube-VOS} & \multicolumn{4}{c}{DAVIS}  \\
\cline{1-9}
Models & PSNR $\uparrow$ & SSIM $\uparrow$ & VFID $\downarrow$ &  $\text{E}_{warp} (\times {10}^{-2}) \downarrow$ & PSNR $\uparrow$ & SSIM $\uparrow$ & VFID $\downarrow$ & $ \text{E}_{warp} (\times {10}^{-2}) \downarrow$ \\
\hline
VINet~\cite{vinet} & 29.20 & 0.9434 & 0.072 & 0.1490 / - & 28.96 & 0.9411 & 0.199 & 0.1785 / -\\
\hline
DFVI~\cite{dfvi}  & 29.16 & 0.9429 & 0.066 & 0.1509 / - & 28.81 & 0.9404 & 0.187 & 0.1880 / 0.1608$^{*}$ \\
\hline
LGTSM~\cite{lgtsm}  & 29.74 & 0.9504 & 0.070 & 0.1859 / - & 28.57 & 0.9409 & 0.170 & 0.2566 / 0.1640$^{*}$ \\
\hline
CAP~\cite{cap}  & 31.58 & 0.9607 & 0.071 & 0.1470 / -   & 30.28 & 0.9521 & 0.182 & 0.1824 / 0.1533$^{*}$ \\
\hline
STTN~\cite{sttn}  & 32.34 & 0.9655 & 0.053 & 0.1451 / 0.0884$^{*}$ & 30.67 & 0.9560 & 0.149 & 0.1779 / 0.1449$^{*}$  \\
\hline\hline

ViB-S & 32.47   & 0.9635  & 0.056  & \qquad\enspace - / 0.0889$^{*}$ &  31.50 & 0.9636 & 0.144 & \qquad\enspace - / 0.1346$^{*}$\\
\hline
ViF & \textbf{33.16}  &  \textbf{0.9673} & \textbf{0.051} & \qquad\enspace - / \textbf{0.0875}$^{*}$ & \textbf{32.54} & \textbf{0.9700} & \textbf{0.138} & \qquad\enspace - / \textbf{0.1336}$^{*}$ \\
\hline
\end{tabular}
\end{center}
\caption{\label{table:completion}Quantitative results of video completion on YouTube-VOS and DAVIS dataset. $^{*}$: our evaluation results following descriptions in STTN~\cite{sttn}, the numerical differences may result from different optical flow models in the evaluation process.}

\end{table*}

\noindent \textbf{Evaluation metrics.}
First, we take Video-based Fr\'echet Inception Distance (\textit{VFID}) as our metric for scoring the perceptually visual quality by comparing with natural video sequences~\cite{wang2018vid2vid, sttn}.
Lower value suggests better realism and visually closer to natural videos.
We also use a optical flow-based warping error $E_{warp}$ for measuring the temporal consistency~\cite{Lai-ECCV-2018}.
Lower value indicates better temporal consistency.
Finally, we use two popular metrics for measuring the quality of reconstructed image compared with original one: Structural SIMilarity (\textit{SSIM}) and Peak Signal to Noise Ratio (\textit{PSNR}). The score is calculated frame by frame and their mean value is reported.
Higher value of these two metrics indicates better reconstruction quality.

\subsection{Ablations}
\paragraph{The effectiveness of soft split and soft composition.}
In Tab. \ref{tab:kernelsize} we show the performance under different patch size used in soft split and soft composition operation on our baseline model ViB-T and ViB-S.
For ViB-T, we keep the stride the same as the patch size. For ViB-S and TiF, they share the same stride $3$ to ensure the same number of tokens for each frames.

First, by changing the patch size for ViB-T, we find that ViB-T with patch size $3$, a straight-forward variant of Transformer has already achieved competitive performance compared to the state-of-the-art STTN \cite{sttn}, even without soft split and soft composition operations.
For ViB-S and ViF, when patch size is larger than $3$, SS and SC operations are incorporated to handle overlap area between patches. All larger patches improves the performance for a significant margin, showing the effectiveness of overlapping patches.
Here we further vary the patch size between SS and SC, limiting the overlapping area to appear in either SS or SC operations. Apart from SS, the overlapped composition in SC can also improve the performance even without SS.

\begin{figure}[t]
    \centering
    \includegraphics[width=0.48\textwidth]{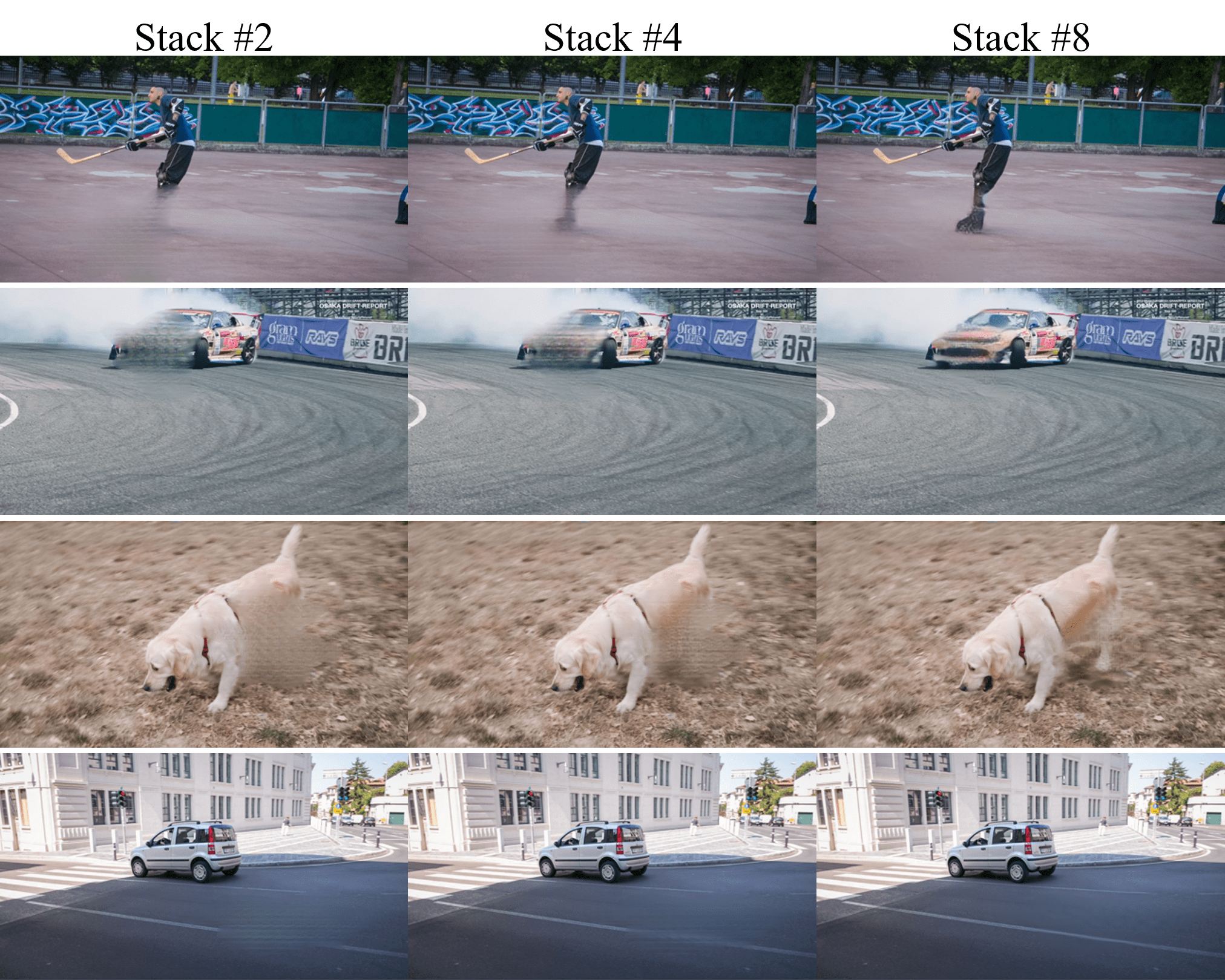}
    \caption{Image decoded from different layers of our trained ViF. It shows that images are refined in a coarse to fine manner.}
   \label{fig:stackablation}
\end{figure}

\begin{figure}[t]
    \centering
    \includegraphics[width=0.48\textwidth]{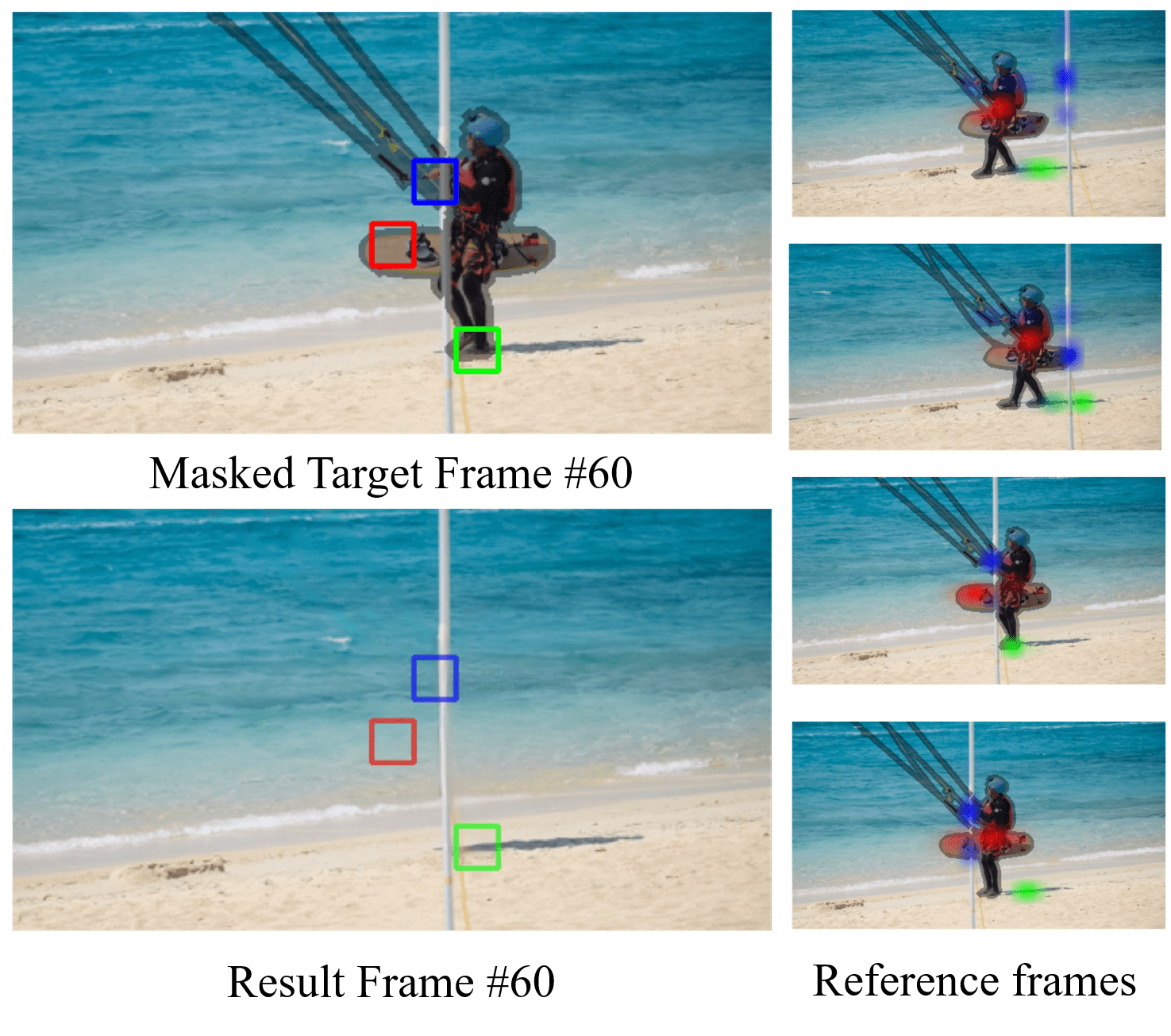}
    \caption{Visualization of attention between patches cross multiple frames in object removal.}
    \label{fig:attention_vis}
\end{figure}

\paragraph{The effectiveness of F3N in FuseFormer.}
As shown in Tab.\ref{tab:kernelsize}, by replacing standard Transformer block with our proposed FuseFormer block in ViB-S, the performance is boosted significantly, showing the effectiveness of sub-token level feature fusion. Moreover, with the proposed normalizing technique in Equ.\ref{equ:meanfold}, the performance has been further improved. Compared to standard Transformer in video inpainting, FuseFormer has slightly fewer parameters and negligible time cost but enabled the sub-token level fine-grain feature fusion.

Fig.\ref{fig:f3n_vis} further illustrates the qualitative results of VIB-S and ViF, demonstrating that their better performance comes from more detailed inpainting results, showing the effectiveness of sub-token level feature fusion.

\subsection{Comparison with other methods}

\noindent \textbf{Qualitative comparison}. In Fig.\ref{fig:final_vis} we show the qualitative results of our model compared with state-of-the-art methods including CAP~\cite{cap}, LGTSM~\cite{lgtsm}, and STTN~\cite{sttn} and our proposed FuseFormer synthesize the most realistic and temporally-coherent videos.

\noindent \textbf{Quantitative comparison}. In Tab.\ref{table:completion} we show the performance comparison with state-of-the-art models on video completion, evaluated on both YouTubeVOS. Our ViF model outperforms all the state-of-the-art video inpainting approaches in video restoration by improving PSNR and SSIM by $3.3\%$ and $0.7\%$, and it yields videos with best realism and temporal coherence by reducing VFID and warping error by $7.4\%$ and $7.8\%$.

\begin{figure}[t]
    \centering
    \includegraphics[width=0.48\textwidth]{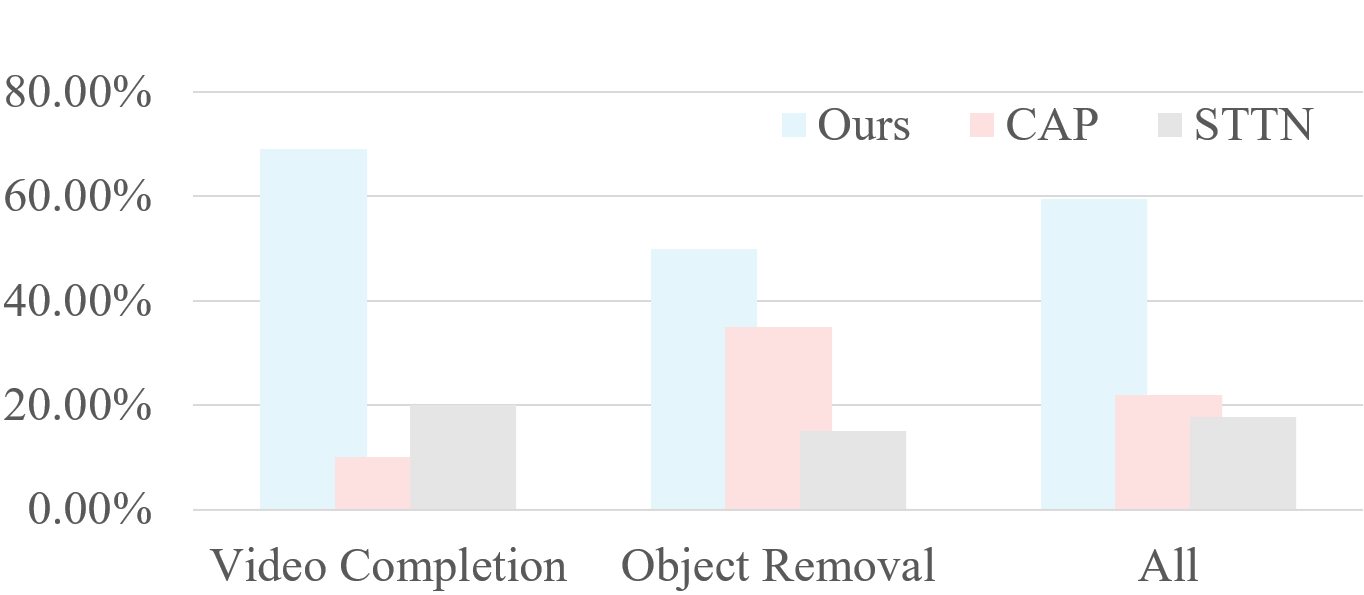}
    \caption{User study results. Percentage of ranking first among 38 viewers of 30 videos on video completion and object removal task.}
    \label{fig:user_study}
\end{figure}
\noindent \textbf{User study}. We choose CAP~\cite{cap} and STTN~\cite{sttn}, two of the state-of-the-art video inpainting models as our baselines for user study. 30 videos are randomly sampled from DAVIS~\cite{Perazzi2016} for object removal and video completion evaluation. 38 volunteers has participated this user study. Videos processed by 3 models are presented at each time for volunteers to rank the inpainting quality.
On our dedicated software for this user study, volunteers can stop/replay any video until they make final judgement. The percentage of first ranking model from each user on each video are shown in Fig.\ref{fig:user_study}, where for both object removal and video completion we have the best performance.

\noindent \textbf{Visualizing inpainting process}. Fig.\ref{fig:stackablation} demonstrates images decoded at different layer of ViF, showing the process of how the our model inpaints a video frame. We can see it starts with coarse context information and gradually refine features in deeper layers.
In Fig.\ref{fig:attention_vis}, we further show the detailed attention process between different multi-frame patches in an object removal task. We can see how our proposed model accurately find reference patch and explore the spatiotemporal information to inpaint the background as well as the pillar.

\section{Conclusion}
In this work we propose FuseFormer, a Transformer model designed for video inpainting via fine-grained feature fusion. It aims at tackling the drawbacks of lacking fine-grained information in patch-based Transformer models. The soft split divides feature map into many patches with given overlapping interval while the soft composition stitches them back into a whole feature map where pixels in overlapping regions are summed up. FuseFormer elaborately builds soft composition and soft split into its feed-forward network for further enhancing sub-patch level feature fusion. Together with our strong Transformer baseline, our FuseFormer model achieve state-of-the-art performance in video restoration and object removal.

\noindent \textbf{Acknowledgement}. This work is supported in part by the General Research Fund through the Research Grants Council of Hong Kong under Grants (Nos. 14204021, 14208417, 14207319, 14202217, 14203118, 14208619), in part by Research Impact Fund Grant No. R5001-18, in part by CUHK Strategic Fund.

{\small
\bibliographystyle{ieee_fullname}

\begin{thebibliography}{10}\itemsep=-1pt

\bibitem{patchmatch}
Connelly Barnes, Eli Shechtman, Adam Finkelstein, and Dan~B Goldman.
\newblock {PatchMatch}: A randomized correspondence algorithm for structural
  image editing.
\newblock {\em ACM Transactions on Graphics (Proc. SIGGRAPH)}, 2009.

\bibitem{10.1145/344779.344972}
Marcelo Bertalmio, Guillermo Sapiro, Vincent Caselles, and Coloma Ballester.
\newblock Image inpainting.
\newblock In {\em Proceedings of the 27th Annual Conference on Computer
  Graphics and Interactive Techniques}, page 417¨C424, 2000.

\bibitem{10.1109/TIP.2003.815261}
M. Bertalmio, L. Vese, G. Sapiro, and S. Osher.
\newblock Simultaneous structure and texture image inpainting.
\newblock {\em IEEE Transactions on Image Processing}, page 882¨C889, 2003.

\bibitem{chang2019free}
Ya-Liang Chang, Zhe~Yu Liu, Kuan-Ying Lee, and Winston Hsu.
\newblock Free-form video inpainting with 3d gated convolution and temporal
  patchgan.
\newblock {\em In Proceedings of the International Conference on Computer
  Vision (ICCV)}, 2019.

\bibitem{lgtsm}
Ya-Liang Chang, Zhe~Yu Liu, Kuan-Ying Lee, and Winston Hsu.
\newblock Learnable gated temporal shift module for deep video inpainting.
\newblock In {\em BMVC}, 2019.

\bibitem{ImageMelding12}
Soheil Darabi, Eli Shechtman, Connelly Barnes, Dan~B Goldman, and Pradeep Sen.
\newblock {I}mage {M}elding: Combining inconsistent images using patch-based
  synthesis.
\newblock {\em ACM Transactions on Graphics (TOG) (Proceedings of SIGGRAPH
  2012)}, 2012.

\bibitem{devlin2018pretraining}
Jacob Devlin, Ming-Wei Chang, Kenton Lee, and Kristina Toutanova.
\newblock Bert: Pre-training of deep bidirectional transformers for language
  understanding, 2018.

\bibitem{vit}
Alexey Dosovitskiy, Lucas Beyer, Alexander Kolesnikov, Dirk Weissenborn,
  Xiaohua Zhai, Thomas Unterthiner, Mostafa Dehghani, Matthias Minderer, Georg
  Heigold, Sylvain Gelly, Jakob Uszkoreit, and Neil Houlsby.
\newblock An image is worth 16x16 words: Transformers for image recognition at
  scale.
\newblock In {\em International Conference on Learning Representations}, 2021.

\bibitem{Efros99texturesynthesis}
Alexei Efros and Thomas Leung.
\newblock Texture synthesis by non-parametric sampling.
\newblock In {\em In International Conference on Computer Vision}, pages
  1033--1038, 1999.

\bibitem{10.1145/383259.383296}
Alexei~A. Efros and William~T. Freeman.
\newblock Image quilting for texture synthesis and transfer.
\newblock In {\em Proceedings of SIGGRAPH}, page 341¨C346.

\bibitem{SMCA}
Peng Gao, Minghang Zheng, Xiaogang Wang, Jifeng Dai, and Hongsheng Li.
\newblock Fast convergence of {DETR} with spatially modulated co-attention.
\newblock {\em CoRR}, abs/2101.07448, 2021.

\bibitem{gan}
Ian~J. Goodfellow, Jean Pouget-Abadie, Mehdi Mirza, Bing Xu, David
  Warde-Farley, Sherjil Ozair, Aaron Courville, and Yoshua Bengio.
\newblock 2014.

\bibitem{He2015}
Kaiming He, Xiangyu Zhang, Shaoqing Ren, and Jian Sun.
\newblock Deep residual learning for image recognition.
\newblock {\em arXiv preprint arXiv:1512.03385}, 2015.

\bibitem{hu2020proposal}
Yuan-Ting Hu, Heng Wang, Nicolas Ballas, Kristen Grauman, and Alexander~G.
  Schwing.
\newblock Proposal-based video completion.
\newblock In {\em The Proceedings of the European Conference on Computer Vision
  (ECCV)}, 2020.

\bibitem{huang2016videocompletion}
Jia-Bin Huang, Sing~Bing Kang, Narendra Ahuja, and Johannes Kopf.
\newblock Temporally coherent completion of dynamic video.
\newblock {\em ACM Trans. Graph.}, 2016.

\bibitem{hudson2021gansformer}
Drew~A Hudson and C.~Lawrence Zitnick.
\newblock Generative adversarial transformers.
\newblock {\em arXiv preprint:2103.01209}, 2021.

\bibitem{IizukaSIGGRAPH2017}
Satoshi Iizuka, Edgar Simo-Serra, and Hiroshi Ishikawa.
\newblock {Globally and Locally Consistent Image Completion}.
\newblock {\em ACM Transactions on Graphics (Proc. of SIGGRAPH)}, 36, 2017.

\bibitem{jiang2021transgan}
Yifan Jiang, Shiyu Chang, and Zhangyang Wang.
\newblock Transgan: Two transformers can make one strong gan.
\newblock {\em arXiv preprint arXiv:2102.07074}, 2021.

\bibitem{vinet}
Dahun Kim, Sanghyun Woo, Joon-Young Lee, and In~So Kweon.
\newblock Deep video inpainting.
\newblock In {\em Proceedings of the IEEE conference on computer vision and
  pattern recognition}, 2019.

\bibitem{adam}
Diederik~P. Kingma and Jimmy Ba.
\newblock Adam: {A} method for stochastic optimization.
\newblock In {\em 3rd International Conference on Learning Representations,
  {ICLR}}, 2015.

\bibitem{alexnet}
Alex Krizhevsky, Ilya Sutskever, and Geoffrey~E. Hinton.
\newblock Imagenet classification with deep convolutional neural networks.
\newblock In {\em Advances in Neural Information Processing Systems 25}. 2012.

\bibitem{Lai-ECCV-2018}
Wei-Sheng Lai, Jia-Bin Huang, Oliver Wang, Eli Shechtman, Ersin Yumer, and
  Ming-Hsuan Yang.
\newblock Learning blind video temporal consistency.
\newblock In {\em European Conference on Computer Vision}, 2018.

\bibitem{cap}
Sungho Lee, Seoung~Wug Oh, DaeYeun Won, and Seon~Joo Kim.
\newblock Copy-and-paste networks for deep video inpainting.
\newblock In {\em Proceedings of the IEEE International Conference on Computer
  Vision}, 2019.

\bibitem{shortlongterm}
Ang Li, Shanshan Zhao, Xingjun Ma, Mingming Gong, Jianzhong Qi, Rui Zhang,
  Dacheng Tao, and Ramamohanarao Kotagiri.
\newblock Short-term and long-term context aggregation network for video
  inpainting.
\newblock In {\em ECCV}, 2020.

\bibitem{liu2018partialinpainting}
Guilin Liu, Fitsum~A. Reda, Kevin~J. Shih, Ting-Chun Wang, Andrew Tao, and
  Bryan Catanzaro.
\newblock Image inpainting for irregular holes using partial convolutions.
\newblock In {\em The European Conference on Computer Vision (ECCV)}, 2018.

\bibitem{liu2019roberta}
Yinhan Liu, Myle Ott, Naman Goyal, Jingfei Du, Mandar Joshi, Danqi Chen, Omer
  Levy, Mike Lewis, Luke Zettlemoyer, and Veselin Stoyanov.
\newblock Roberta: A robustly optimized bert pretraining approach, 2019.

\bibitem{Newson2014:SIIMS-VideoInpainting}
Alasdair Newson, Andr{\'{e}}s Almansa, Matthieu Fradet, Yann Gousseau, and
  Patrick P{\'{e}}rez.
\newblock {Video Inpainting of Complex Scenes}.
\newblock {\em SIAM Journal on Imaging Sciences}, pages 1993--2019, 2014.

\bibitem{detr}
Gabriel Synnaeve Nicolas Usunier Alexander Kirillov Sergey~Zagoruyko
  Nicolas~Carion, Francisco~Massa.
\newblock End-to-end object detection with transformers.
\newblock In {\em European Conference on Computer Vision}, 2020.

\bibitem{opn}
Seoung~Wug Oh, Sungho Lee, Joon-Young Lee, and Seon~Joo Kim.
\newblock Onion-peel networks for deep video completion.
\newblock In {\em Proceedings of the IEEE International Conference on Computer
  Vision}, 2019.

\bibitem{pathakCVPR16context}
Deepak Pathak, Philipp Kr\"ahenb\"uhl, Jeff Donahue, Trevor Darrell, and Alexei
  Efros.
\newblock Context encoders: Feature learning by inpainting.
\newblock In {\em Computer Vision and Pattern Recognition ({CVPR})}, 2016.

\bibitem{Perazzi2016}
F. Perazzi, J. Pont-Tuset, B. McWilliams, L. {Van Gool}, M. Gross, and A.
  Sorkine-Hornung.
\newblock A benchmark dataset and evaluation methodology for video object
  segmentation.
\newblock In {\em Computer Vision and Pattern Recognition}, 2016.

\bibitem{nlppre}
Alec Radford and Ilya Sutskever.
\newblock Improving language understanding by generative pre-training.
\newblock 2018.

\bibitem{Strobel2014FlowAC}
M. Strobel, Julia Diebold, and D. Cremers.
\newblock Flow and color inpainting for video completion.
\newblock In {\em GCPR}, 2014.

\bibitem{attention}
Ashish Vaswani, Noam Shazeer, Niki Parmar, Jakob Uszkoreit, Llion Jones,
  Aidan~N Gomez, \L~ukasz Kaiser, and Illia Polosukhin.
\newblock Attention is all you need.
\newblock In {\em Advances in Neural Information Processing Systems}, pages
  5998--6008, 2017.

\bibitem{wang2019aaai}
Chuan Wang, Haibin Huang, Xiaoguang Han, and Jue Wang.
\newblock Video inpainting by jointly learning temporal structure and spatial
  details.
\newblock In {\em AAAI}, 2019.

\bibitem{wang2018vid2vid}
Ting-Chun Wang, Ming-Yu Liu, Jun-Yan Zhu, Guilin Liu, Andrew Tao, Jan Kautz,
  and Bryan Catanzaro.
\newblock Video-to-video synthesis.
\newblock In {\em Advances in Neural Information Processing Systems (NeurIPS)},
  2018.

\bibitem{wang2020end}
Yuqing Wang, Zhaoliang Xu, Xinlong Wang, Chunhua Shen, Baoshan Cheng, Hao Shen,
  and Huaxia Xia.
\newblock End-to-end video instance segmentation with transformers.
\newblock In {\em Proc. IEEE Conf. Computer Vision and Pattern Recognition
  (CVPR)}, 2021.

\bibitem{youtubevos}
Ning Xu, Linjie Yang, Yuchen Fan, Dingcheng Yue, Yuchen Liang, Jianchao Yang,
  and Thomas Huang.
\newblock Youtube-vos: A large-scale video object segmentation benchmark.
\newblock {\em arXiv: 1809.03327}, 2018.

\bibitem{dfvi}
Rui Xu, Xiaoxiao Li, Bolei Zhou, and Chen~Change Loy.
\newblock Deep flow-guided video inpainting.
\newblock In {\em Proceedings of the IEEE conference on computer vision and
  pattern recognition}, 2019.

\bibitem{yu2018free}
Jiahui Yu, Zhe Lin, Jimei Yang, Xiaohui Shen, Xin Lu, and Thomas~S Huang.
\newblock Free-form image inpainting with gated convolution.
\newblock {\em arXiv preprint arXiv:1806.03589}, 2018.

\bibitem{yu2018generative}
Jiahui Yu, Zhe Lin, Jimei Yang, Xiaohui Shen, Xin Lu, and Thomas~S Huang.
\newblock Generative image inpainting with contextual attention.
\newblock In {\em Proceedings of the IEEE conference on computer vision and
  pattern recognition}, pages 5505--5514, 2018.

\bibitem{yuan2021tokens}
Li Yuan, Yunpeng Chen, Tao Wang, Weihao Yu, Yujun Shi, Francis~EH Tay, Jiashi
  Feng, and Shuicheng Yan.
\newblock Tokens-to-token vit: Training vision transformers from scratch on
  imagenet.
\newblock {\em arXiv preprint arXiv:2101.11986}, 2021.

\bibitem{sttn}
Yanhong Zeng, Jianlong Fu, and Hongyang Chao.
\newblock Learning joint spatial-temporal transformations for video inpainting.
\newblock In {\em The Proceedings of the European Conference on Computer Vision
  (ECCV)}, 2020.

\bibitem{zhang2019internal}
Haotian Zhang, Long Mai, Ning Xu, Zhaowen Wang, John Collomosse, and Hailin
  Jin.
\newblock An internal learning approach to video inpainting.
\newblock In {\em Proceedings of the IEEE International Conference on Computer
  Vision}, pages 2720--2729, 2019.

\bibitem{transcluster}
Minghang Zheng, Peng Gao, Xiaogang Wang, Hongsheng Li, and Hao Dong.
\newblock End-to-end object detection with adaptive clustering transformer.
\newblock {\em CoRR}, abs/2011.09315, 2020.

\bibitem{zhou2018end}
Luowei Zhou, Yingbo Zhou, Jason~J Corso, Richard Socher, and Caiming Xiong.
\newblock End-to-end dense video captioning with masked transformer.
\newblock In {\em Proceedings of the IEEE Conference on Computer Vision and
  Pattern Recognition}, pages 8739--8748, 2018.

\bibitem{zhu2020deformable}
Xizhou Zhu, Weijie Su, Lewei Lu, Bin Li, Xiaogang Wang, and Jifeng Dai.
\newblock Deformable detr: Deformable transformers for end-to-end object
  detection.
\newblock {\em arXiv preprint arXiv:2010.04159}, 2020.

\end{thebibliography}

}

\end{document}